%% file: main.tex
\documentclass{egpubl}
\usepackage{pg2024}

\SpecialIssueSubmission    

\usepackage[T1]{fontenc}
\usepackage{dfadobe}  

\usepackage{cite}  

\usepackage{float}
\usepackage{calc}
\usepackage{subfigure}
\usepackage{multirow}
\usepackage{amsmath}
\usepackage{amssymb}
\usepackage{algorithm}
\usepackage{algorithmic}
\usepackage{bbm}
\usepackage{amsfonts}
\usepackage{mathrsfs}
\usepackage{capt-of,etoolbox}
\usepackage{booktabs}
\usepackage{xcolor}
\BibtexOrBiblatex
\electronicVersion
\PrintedOrElectronic
\ifpdf \usepackage[pdftex]{graphicx} \pdfcompresslevel=9
\else \usepackage[dvips]{graphicx} \fi

\usepackage{egweblnk} 

\newcommand{\hang}[1]{{\color{black}#1}}

\newcommand{\hangz}[1]{{\color{black}#1}}

\title[DiffPop: Plausibility-Guided Object Placement Diffusion for Image Composition]%
      {DiffPop: Plausibility-Guided Object Placement Diffusion for \\ Image Composition}

\author[J.C. Liu et al.]
{\parbox{\textwidth}{\centering Jiacheng Liu$^{1}$, Hang Zhou$^{2}$, Shida Wei$^{1}$, Rui Ma\thanks{Corresponding author}$^{1}$}
        \\
{\parbox{\textwidth}{\centering $^1$Jilin University, China\\
         $^2$Simon Fraser University, Canada
       }
}
}

\begin{document}

\teaser{
 \includegraphics[width=\linewidth]{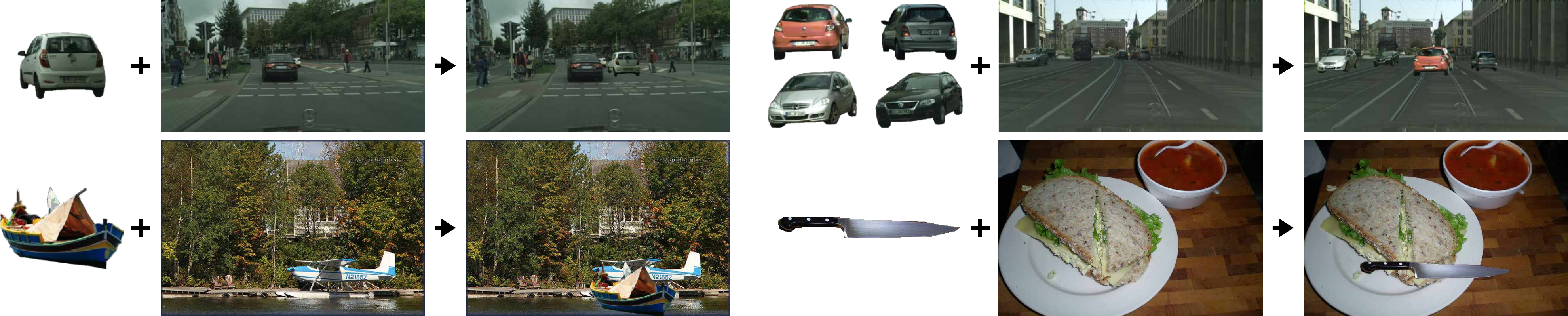}
 \centering
  \caption{
  Given one or multiple foreground objects, our plausibility-guided diffusion model can generate plausible object placement with diverse scale and location, as well as structural coherence to the background image.
  }
\label{img_new_1}
}

\maketitle

\input{00-abstract}
\input{01-intro}
\input{02-related-work}
\input{03-methods}
\input{04-experiments}
\input{05-conclusion}

\bibliographystyle{eg-alpha-doi} 
\bibliography{main}   

\input{06-Supplementary-material}


\newpage

\end{document}

%% file: 00-abstract.tex
\begin{abstract}
In this paper, we address the problem of plausible object placement for the challenging task of realistic image composition. 
We propose DiffPop, the first framework that utilizes plausibility-guided denoising diffusion probabilistic model to learn the scale and spatial relations among multiple objects and the corresponding scene image.
First, we train an unguided diffusion model to directly learn the object placement parameters in a self-supervised manner.
Then, we develop a human-in-the-loop pipeline which exploits human labeling on the diffusion-generated composite images to provide the weak supervision for training a structural plausibility classifier. 
The classifier is further used to guide the diffusion sampling process towards generating the plausible object placement.
Experimental results verify the superiority of our method for producing plausible and diverse composite images on the new Cityscapes-OP dataset and the public OPA dataset, as well as demonstrate its potential in applications such as data augmentation and multi-object placement tasks.
Our dataset and code will be released. 

\ccsdesc[500]{Computing methodologies~Image manipulation}
\ccsdesc[500]{Computing methodologies~Computer vision}

\printccsdesc   
\end{abstract}  

%% file: 01-intro.tex
\section{Introduction}\label{sec1}

Image composition involves creating realistic composite images by combining specific foreground objects with background images. It has wide applications in fields such as entertainment and creative industries. In this work, we focus on the object placement task, which is a subtask of image composition. Object placement refers to accurately pasting a foreground object onto a background image with a suitable scale and location, resulting in a realistic composite image. This technology finds utility in various scenarios. For example, in product or advertisement design, object placement can provide suggestions for the optimal placement of logos. 
In entertainment content, such as popular mobile phone AR applications, virtual objects can be automatically inserted into scenes for visualization or interaction.

While several works \cite{context,st-gan,placenet,terse,sac,graph,topnet} have been proposed to address the object placement task, there are still challenges to be addressed. Previous methods \cite{context,st-gan,placenet,graph} adopt generative adversarial networks (GANs) \cite{GAN} for adversarial training, which are prone to mode collapse and lack of diversity in generation. To mitigate such issue, more stable generators such as VAE-GANs \cite{vae-gan,sac} are adopted later, yet mode collapse issue still exists. Another challenge lies in the lack of training data for complex scenes. 
Taking Cityscapes dataset \cite{cityscapes} as an example, the existing object scales and locations are not sufficient for training a functional placement network. 
The main reason is, training solely on real images leads to an abundance of positive samples but no negative samples, resulting in lower plausibility of the composite image. 
Meanwhile, the recent OPA dataset \cite{opa} provide positive and negative binary labels for rational or plausible object placement.
With these labels, the OPA dataset is suitable for training classifiers for \textit{assessing} the binary plausibility of the object placement result.
However, how to efficiently utilize OPA's positive and negative labels for \textit{generating} plausible object placement still remains to be a problem.

In contrast to GANs which are prone to unstable training and limited diversity, diffusion models \cite{ddpm,ddim,iddpm,beat-gan,class-free} provide the advantage of stable training and enhanced diversity. 
By continuously adding noise to real samples and then denoising, diffusion models enable the generation of realistic samples and demonstrate state-of-the-art performance in various image generation tasks, including unconditional image generation, image inpainting, and image super-resolution.
Compared to other generative models, diffusion models can capture and model the intricate dependencies in images and lead to visually impressive results and improved generation performance. 
Furthermore, the classifier-guided diffusion models \cite{beat-gan} can utilize the classifier guidance during the sampling process to enhance the generation quality of specific class samples.
It is a natural thought to explore the applicability of diffusion models for the object placement task.

In this paper, we propose DiffPop, the first \textbf{p}lausibility-guided \textbf{diff}usion probabilistic model for \textbf{o}bject \textbf{p}lacement. 
Initially, we employ a self-supervised training scheme to train a object placement diffusion model directly on the object transformation parameters, without explicit conditioning on the foreground and background information. 
This enables the diffusion model to learn the distribution of scales and locations for foreground objects w.r.t. background scenes from real data.
Then, we train a structural plausibility classifier which evaluates the generated placement at each time step and guides the diffusion sampling process towards the desired direction, i.e., plausible object placement.
Specifically, when training such a plausibility classifier, we either take the existing positive/negative labels from the existing dataset (e.g., OPA) or adopt a human-in-the-loop strategy to address the issue of plausibility measurement.
For the latter case, we manually label the composite images produced from our initial unguided diffusion model into distinct positive and negative classes, based on the criterion of the image-level plausibility and realism.
Such easy-to-obtain human annotations can provide sufficient weak supervision for learning the binary plausibility classifier, which can be efficiently used in the guided object placement diffusion.

To verify the effectiveness of our method, we conduct extensive experiments on the OPA dataset and Cityscapes-OP, a new dataset created by manually labeling the composite images produced from the initial unguided diffusion model.
In addition, we also show the applications of our method in creating composite images for data augmentation, as well as its extension for multi-object placement.

In summary, our contributions are as follows: 
\begin{itemize}
\item We propose DiffPop, the first plausibility-guided diffusion framework that aims to generate plausible object placement for image composition. 
Specifically, we learn a structural plausibility classifier to provide guidance on the diffusion-based object placement generation process.
\item We employ the human-in-the-loop strategy to obtain image-level weak supervision for training the plausibility classifier. 
And we create a new dataset Cityscapes-OP, which can be used for training plausibility-guided diffusion model for placing objects on scenes with more complex and structural backgrounds than those in OPA dataset.
\item Experimental results demonstrate that our method achieves state-of-the-art object placement performance in terms of plausibility and diversity on both Cityscapes-OP and OPA datasets. 
Our approach also shows promising results in creating composite images for data augmentation and multi-object placement.
\end{itemize}

%% file: 02-related-work.tex
\section{Related work}\label{sec2}

\subsection{Object placement}\label{subsec1}

Image composition involves creating a composite image that appears realistic by combining a specific foreground object with a background image. In the field of computer vision, Niu et al. \cite{survey} categorized image composition into four branches: object placement, image blending, image harmonization, and shadow generation. These branches address various challenges encountered during the image composition process. 
In this paper, we focus on the object placement task which aims to determining the appropriate scale and location for the foreground object. 

Traditional object placement methods \cite{o1,o2,o3,o4,o5} employ explicit rules to find suitable locations and scales for the foreground object. On the other hand, learning-based object placement methods \cite{sac,context,graph,placenet,terse} typically predict or generate affine transformation matrices to determine the location and scale of the foreground object on the background image. Lin et al. \cite{st-gan} introduced a novel GAN architecture that leverages a spatial transformer network (STN) as a generator to transform the foreground objects based on generated transformation parameters, resulting in the generation of realistic composite images. 
This deep learning-based approach has significantly advanced the field of object placement.
Lee et al. \cite{context} proposed an end-to-end VAE-GAN that generates transformation matrices and shapes for objects through self-supervised and unsupervised training. This method mitigates the risk of mode collapse during GAN training and has been widely adopted in subsequent work. Tripathi et al. \cite{terse} incorporated an additional discriminator network during GAN training to facilitate targeted data augmentation for downstream tasks. Zhang et al. \cite{placenet} utilized self-supervised data pairs obtained through pre-trained instance segmentation and image inpainting methods to ensure diversity for object placement. Liu et al. \cite{opa} created a dedicated object placement dataset called OPA and introduced the SimOPA classifier to assess the object placement. Zhou et al. \cite{graph} transformed the object placement problem into a graph node completion task and employed binary classification loss to train the discriminator network, which makes full use of labeled negative samples. SAC-GAN \cite{sac} incorporated edge and semantic information from the object and background image to improve the structural coherence of the composite results. TopNet \cite{topnet} proposed the use of Transformers to learn the relationship between object features and local background features, resulting in improved generation of object scale and location.

In contrast to the aforementioned methods, our approach is based on the diffusion model, which offers diversity and stable training. Our method effectively utilizes positive and negative samples to guide the diffusion model in generating more reasonable scales and locations, leading to the composition of realistic images. 
Additionally, our guided-diffusion framework can also be extended to simultaneously placement of multiple objects, which cannot be achieved by previous methods.

\subsection{Diffusion models}\label{subsec2}

Ho et al. \cite{ddpm} introduced denoising diffusion probabilistic models (DDPM), a generative model that optimizes the network by continuously adding noise to real samples and using the network to denoise. This approach enables the network to generate realistic samples. Song et al. \cite{ddim} made improvements to the diffusion model to enhance sampling speed. Nichol et al. \cite{iddpm} further enhanced the diffusion model's ability to generate high-quality samples. 
For conditional image synthesis, Dharival et al. \cite{beat-gan} enhanced sampling quality by incorporating classifier guidance during the diffusion model's sampling process. They utilized the gradient of the classifier to balance the diversity and plausibility of the generated samples. Liu et al. \cite{more-control} introduced a unified semantic diffusion guidance framework that allows guidance through language, image, or both. Ho et al. \cite{class-free} jointly trained conditional and unconditional diffusion models, combining the resulting conditional and unconditional score estimates to achieve a balance between sample quality and diversity. This approach frees the diffusion model from the limitations of classifier-guided sampling.
Nichol et al. \cite{glide} proposed GLIDE, a model capable of generating high-quality images conditioned on text. They demonstrated that unconditional guidance is superior to CLIP guidance for language-based conditioning. Ramesh et al. \cite{dalle2} proposed unCLIP, which leverages the feature space of CLIP and the diffusion model to generate images from textual descriptions in a zero-shot manner. Saharia et al. \cite{imagen} introduced Imagen, a framework that combines large Transformer language models with diffusion models to empower the network with the ability to generate images from textual prompts. Robin et al. \cite{stabel-diffusion} applied diffusion models directly to latent spaces, resulting in significant computational resource savings for text-to-image generation. Hachnochi et al. \cite{image-compos-diffuion} applied diffusion models to the field of image composition. They iteratively injected contextual information from background images into inserted foreground objects, allowing control over the degree of change in the foreground object. 

In contrast to above methods, we focus on the plausibility-guided object placement, i.e., we first train a plausibility-classifier based on the weakly annotated images produced from an unguided diffusion model and then use the classifier to guide the diffusion sampling process for producing plausible results.


%% file: 03-methods.tex
\begin{figure*}[ht]
\centering
\includegraphics[width=\textwidth]{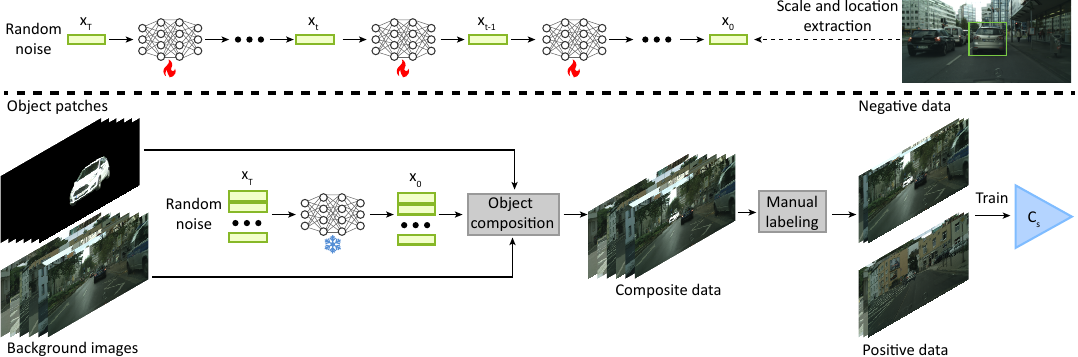} 
\caption{
The training pipeline of our DiffPop, consisting of two stages. In Stage-1 (above), we train an unguided object placement denoising diffusion model on object scales and locations. In Stage-2 (below), we first utilize the generated scales and locations from above pre-trained diffusion model to form the corresponding transformation matrix, followed by applying object composition to obtain the composite images; then, we adopt the human-in-the-loop strategy and manually label composite images into positive and negative classes based on plausibility and realism of composite images. These labeled data are then used for training the structural plausibility classifier $\mathbf{C}_s$. 
}
\label{img_new_3}
\end{figure*}

\begin{figure*}[ht]
\centering

\includegraphics[width=0.98\textwidth]{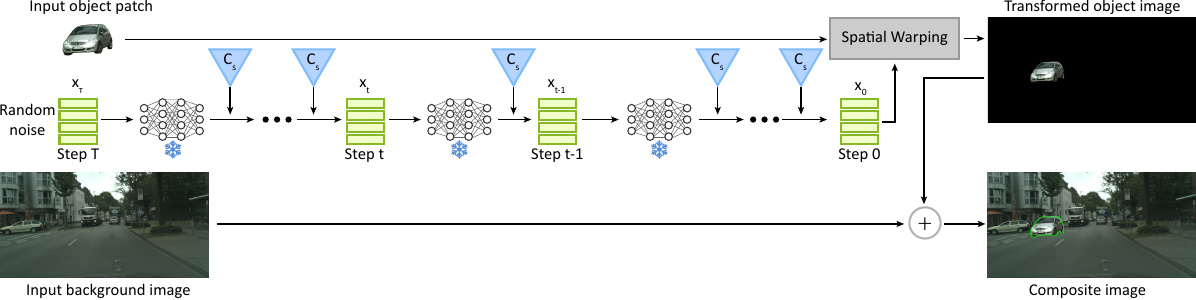} 
\caption{The inference pipeline of our DiffPop. We sample a random placement $\mathbf{x}_T$ from the Gaussian noise at step $T$ and iteratively denoise it till step $0$ to obtain the desired placement $\mathbf{x}_0$. When sampling at step $t$, we take the gradient from the structural plausibility classifier $\mathbf{C}_s$ for guided generation towards scene-level structural coherence. 
}
\label{img_new_0}
\end{figure*}

\section{Method}\label{sec3}

Given a scene image as background and an object patch as foreground,
we seek to learn scale and spatial distributions of object placement so that realistic image composition can be obtained. 
The training pipeline of our DiffPop framework contains two stages as shown in Fig.~\ref{img_new_3}:
in Stage-1, an unguided object placement denoising diffusion model is trained to learn the distributions of object scales and locations;
in Stage-2, a structural plausibility classifier is trained with the manually labeled composite images generated by the previously trained model.
At inference time, as shown in Fig.~\ref{img_new_0}, given a background image and an object patch, we generate 2D transformation (scale and location) for plausible object placement using the classifier-guided diffusion and adopt the copy-paste scheme for image composition. 

\subsection{Unguided object placement denoising diffusion}\label{subsec3}
As described before, we first train an unguided denoising diffusion model for learning the distributions of object scales and locations in the given dataset.

\textbf{Diffusion process.}
The forward diffusion process is a pre-defined Markov chain that operates on object placement $\mathbf{x}\in\mathbb{R}^{D}$, where $\mathbf{x}=[s, v, h]$, $s$ is the relative scale of the object-image pair and $(v, h)$ are the vertical and horizontal offsets of the object to the image, as shown in Fig. \ref{img_new_3} (above). To initiate the diffusion process, we start with a clean object placement $\mathbf{x}_0$ sampled from the underlying distribution $q(\mathbf{x}_0)$. Then, we gradually add Gaussian noise to $\mathbf{x}_0$, resulting in a series of intermediate object placement variables $\mathbf{x}_{1:T}$ following a predetermined schedule of linearly increasing noise variances, denoted as $\beta_{1:T}$. The joint distribution $q\left(\mathbf{x}_{t} | \mathbf{x}_{t-1}\right)$ of the diffusion process is formulated as: 
\begin{equation}
\begin{aligned}
     q\left(\mathbf{x}_{1: T} | \mathbf{x}_{0}\right) = \prod_{t = 1}^{T} q\left(\mathbf{x}_{t} | \mathbf{x}_{t-1}\right), 
\end{aligned}
\end{equation}
where the diffusion step at time $t$ is defined as:
\begin{equation}
\begin{aligned}
     q\left(\mathbf{x}_{t} | \mathbf{x}_{t-1}\right)=\mathcal{N}\left(\mathbf{x}_{t} ; \sqrt{1-\beta_{t}} \mathbf{x}_{t-1}, \beta_{t} \mathbf{I}\right).
\end{aligned}
\end{equation}

Thanks to the properties of Gaussian distribution, we can directly sample $\mathbf{x}_{t}$ without the recursive formulation by:
\begin{equation}
\begin{aligned}
     q\left(\mathbf{x}_{t} | \mathbf{x}_{0}\right)=\mathcal{N}\left(\mathbf{x}_{t} ; \sqrt{\bar{\alpha}_{t}} \mathbf{x}_{0},\left(1-\bar{\alpha}_{t}\right) \mathbf{I}\right),
\end{aligned}
\end{equation}
where $\mathbf{x}_{t}=\sqrt{\bar{\alpha}_t}\mathbf{x}_{0}+\sqrt{1-\bar{\alpha}_t}\epsilon$, $\alpha_t=1-\beta_t$, $\bar{\alpha}_t=\prod_{r=1}^{t}\alpha_t$, $\epsilon$ is the noise to corrupt $\mathbf{x}_{t}$.

\textbf{Denoising process.}
The denoising process, also known as generative process, is parameterized as a Markov chain with learnable reverse Gaussian transitions. Given a noisy object placement sampled from a standard multivariate Gaussian distribution, denoted as $\mathbf{x}_{T}\sim\mathcal{N}(\mathbf{0},\mathbf{I})$, serving as the initial state. The goal is to correct each state $\mathbf{x}_{t}$ at every time step, producing a cleaner version $\mathbf{x}_{t-1}$ using the learned Gaussian transition $p_{\theta}\left(\mathbf{x}_{t-1} | \mathbf{x}_{t}\right)$. This transition is determined by a learnable network denoted as $\Theta$. By iteratively applying this reverse process until the maximum number of steps $T$ is reached, the final state $\mathbf{x}_{0}$, representing the desired clean object placement, is obtained. The joint distribution of the generative process, denoted as $p_{\theta}\left(\mathbf{x}_{0: T}\right)$, is expressed as follows:
\begin{equation}
\begin{aligned}
     p_{\theta}\left(\mathbf{x}_{0: T}\right)=p\left(\mathbf{x}_{T}\right) \prod_{t=1}^{T} p_{\theta}\left(\mathbf{x}_{t-1} | \mathbf{x}_{t}\right),
\end{aligned}
\end{equation}
\begin{equation}
\begin{aligned}
     p_{\theta}\left(\mathbf{x}_{t-1} | \mathbf{x}_{t}\right)=\mathcal{N}\left(\mathbf{x}_{t-1} ; \mu_{\theta}\left(\mathbf{x}_{t}, t\right), \boldsymbol{\Sigma}_{\theta}\left(\mathbf{x}_{t}, t\right)\right),
\end{aligned}
\end{equation}
where the parameters $\mu_{\theta}\left(\mathbf{x}_{t}, t\right)$ and $\boldsymbol{\Sigma}_{\theta}\left(\mathbf{x}_{t}, t\right)$ represent the predicted mean and covariance, respectively, of the Gaussian distribution for $\mathbf{x}_{t-1}$. These parameters are obtained by taking $\mathbf{x}_{t}$ as input into the denoising network $\Theta$. For simplicity, we set predefined constants for $\boldsymbol{\Sigma}_{\theta}\left(\mathbf{x}_{t}, t\right)$ as in DDPM~\cite{ddpm}. Subsequently, $\mu_{\theta}\left(\mathbf{x}_{t}, t\right)$ can be reparameterized by subtracting the predicted noise according to Bayes's theorem:
\begin{equation}
\begin{aligned}
     \mu_{\theta}\left(\mathbf{x}_{t}, t\right)=\frac{1}{\sqrt{\alpha_{t}}}\left(\mathbf{x}_{t}-\frac{\beta_{t}}{\sqrt{1-\bar{\alpha}_{t}}} \epsilon_{\theta}\left(\mathbf{x}_{t}, t\right)\right).
\end{aligned}
\end{equation}

\textbf{Network training objective.}
We extract the ground-truth object placements from the images to train our placement denoising network in a self-supervised manner. The network $\Theta$ is a simply 4-layer MLPs, with input and output sizes of $N\times 3$. We train the network following $\epsilon$-prediction from DDPM~\cite{ddpm} with $\ell_2$ loss: 
\begin{equation}
\begin{aligned}
    L_{\text {unguided}}=\mathbb{E}_{\mathbf{x}_{0}, \epsilon, t}\left\|\epsilon-\epsilon_{\theta}\left(\mathbf{x}_{t}, t\right)\right\|^{2}.
\end{aligned}
\end{equation}

\subsection{Plausibility-guided object placement diffusion}
As the unguided diffusion model only learns the object placement distribution, it does not consider the scene-level structural coherence and may fail to generate plausible placement conditioned on given objects and scene images.
Inspired by the classifier-guided conditional generation in \cite{beat-gan}, we train a classifier based on annotated structural plausibility, and use its gradient to guide the diffusion sampling process towards scene-level structural coherence.
To train such a plausibility classifier, we either take the existing positive/negative labels from the existing dataset (e.g., OPA) or adopt a human-in-the-loop strategy to address the annotation of plausibility measurement.


\textbf{Human-in-the-loop plausibility labeling.}
Existing datasets like Cityscapes were not initially designed for the object placement tasks, resulting in lack of ground-truth annotations for training an object composition network. 
Although self-supervised training scheme \cite{placenet, sac} can be used to learn the object placement distributions from positive examples, the negative examples which are essential for training a binary classifier for plausibility measurement are generally missing.
To resolve this issue, we employ a human-in-the-loop strategy to manually assign positive and negative labels to the composite images produced by the unguided object placement diffusion model, based on the criterion of the plausibility and realism of the images, as shown in Fig. \ref{img_new_3} (below).
Simply, a positive label means it is structure-coherent between the inserted object and the background scene, and the overall image is plausible, and vise versa.
These human annotations can provide essential weak supervision for learning the plausibility classifier which can measure the results from the unguided diffusion model and further be used to guide the diffusion sampling process.

\textbf{Structural plausibility classifier.}
To guide the diffusion model towards generating plausible object placements, we train a structural plausibility classifier $\mathbf{C}_s$ and use its gradient to guide the diffusion sampling process.
The $\mathbf{C}_s$ is simply defined as a ResNet-18 backboned binary classifier, targeting to judge the structural plausibility of the composite image at the scene-level. 
The classifier takes the semantic scene layout combined with the object mask as inputs and trained with manually annotated positive/negative labels in a supervised manner. 
For the semantic layout of input scene, we directly utilize the semantic map provided by the dataset and process it into binary masks, while each mask is corresponding to one category. 
To obtain the composite layout from the input object mask and processed binary scene layout, we transform the 2D object patch using spatial warping proposed by spatial transformer network (STN)~\cite{stn} with the affine transformation matrix $A_t$ generated by the unguided diffusion model, where 
\begin{equation}
\label{eq:affine}
A_t  = \left[\begin{array}{lll}
s_t & 0 & h_t \\
0 & s_t & v_t
\end{array}\right].
\end{equation}%
The classifier is independently trained using results from unguided diffusion and will be frozen during the guided diffusion process.

\begin{figure}[t]
\centering

\includegraphics[width=\columnwidth]{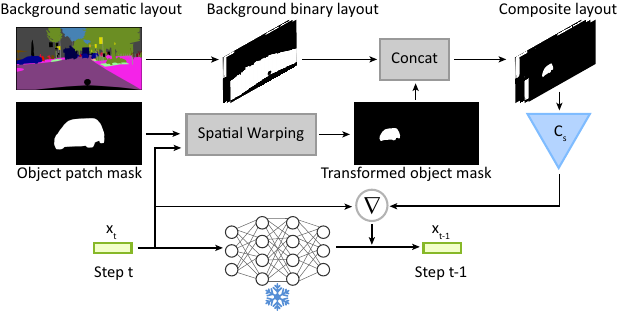} 
\caption{
\hang{
The detailed illustration for biased sampling guided by the structural plausibility classifier $\mathbf{C}_s$. $\mathbf{C}_s$ takes the composite layouts as the input and outputs the probability of structural plausibility, which is further utilized in the guided sampling of object placement diffusion model at step $t-1$. 
}
}
\label{img_new_4}
\end{figure}

\textbf{Classifier-guided diffusion.}
Once the classifier $\mathbf{C}_s$ is trained, we use its gradient to guide the sampling process of the object placement diffusion model (Fig.~\ref{img_new_4}). Specifically, the sampling formula is defined as follows:
\begin{equation}
\begin{aligned}
    \mathbf{x}_{t-1} \leftarrow \mathcal{N}\left(\mu + \lambda \Sigma \nabla_{\mathbf{x}_{t}} \log p_{\phi}(\mathbf{C}_s(\mathbf{x}_t) | \mathbf{x}_{t}), \Sigma\right),
\end{aligned}
\end{equation}
where $\lambda$ is the guidance scale factor which determines how much the gradient of the classifier $\mathbf{C}_s$ affect the sampling of the diffusion model, $\phi$ is the parameters of $\mathbf{C}_s$. 
Specifically, we generate a transformation based on the sampling result of time $t$ and utilize it to transform a given object patch and combine it with the target background layout.
Then, the composite layout (see Sec. \ref{sec:ablation} for details) is fed into $\mathbf{C}_s$ to obtain a plausibility score, which is the probability of the binary classification. 
We compute the gradient of the score w.r.t. the sampling result at time $t$, and the gradient is used to guide the sampling result at time $t-1$.
After $T$ iterations, the final placement $\mathbf{x}_0=[s_{0}, h_{0}, v_{0}]$ can be obtained. 
Finally, we use the affine transformation matrix $A_0$ formed by $(s_{0}, h_{0}, v_{0})$ (see Eq.~\ref{eq:affine}) to transform the object, and paste the transformed object to the background image to obtain the composite image. 

%% file: 04-experiments.tex
\section{Results}\label{sec4}

We conduct a series of experiments to test how DiffPop performs on two datasets: Cityscapes-OP and OPA. 
Quantitative and qualitative comparisons with related methods show the superiority of our method in generating plausible and diverse results. 
In addition, we also explore the potential of our method in applications such as data augmentation and multi-object placement.

\subsection{Dataset}\label{subsec6}

\textbf{Cityscapes-OP dataset.}
We build Cityscapes-OP based on composite images produced from the unguided diffusion model trained on the Cityscapes \cite{cityscapes} dataset.
Specifically, we collect 75 different foreground objects and use the unguided diffusion model to generate the placement of each object onto 600 different background scenes.
The full dataset provides human-annotated plausibility labels for all composite images, including 12,000 (3,869 positive / 8,131 negative) composite images for training and 1,000 (395 positive / 605 negative) composite images for testing. 
More details on creating the dataset can be found in the supplementary material. 

\textbf{OPA dataset.}
OPA \cite{opa} dataset is specifically designed for assessing the object placement. 
Hence, each image is annotated with positive/negative rationality labels indicating whether the object placement is rational (or plausible).
All the images are collected from COCO \cite{coco} dataset and each composite image is generated with one background, one object, and one placement bounding box. 
The dataset includes 62,074 (21,376 positive/40,698 negative) composite images for training and 11,396 (3,588 positive/7,808 negative) composite images for testing. 
The composite images in OPA dataset contain 1,389 different background scenes and 4,137 different foreground objects in 47 different categories. 

\begin{figure*}[t]
\centering
\includegraphics[width=0.98\textwidth]{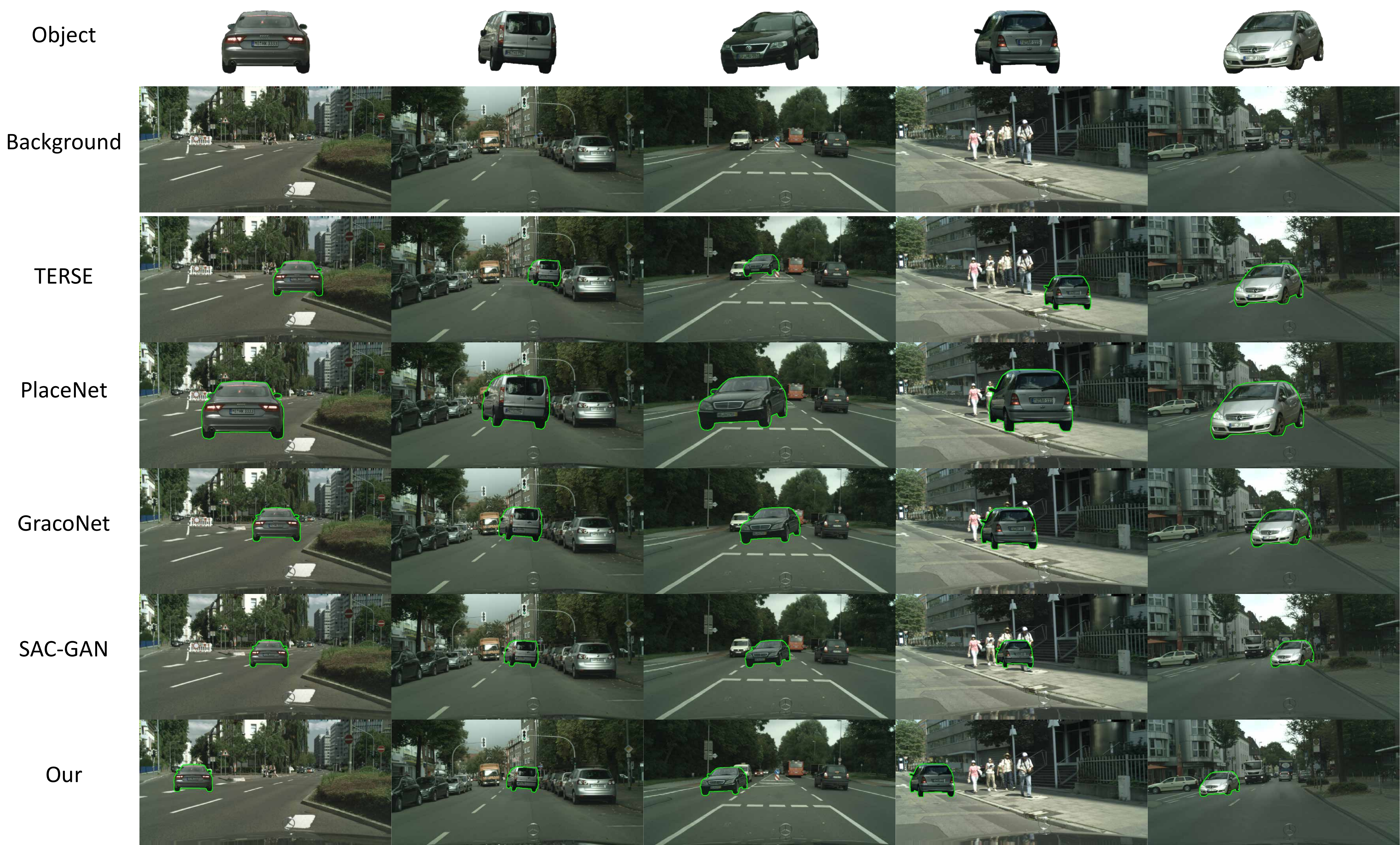} 
\caption{
Qualitative results of single object placement on Cityscapes-OP. Best viewed with zoom-in. 
}
\label{img5}
\end{figure*}

\subsection{Network training details}\label{subsec7}
We trained two networks in our DiffPop framework, namely the diffusion network $\Theta$ and classifier $\mathbf{C}_s$. 
All models are implemented using PyTorch \cite{pytorch} and trained on one RTX 3060 GPU. 
By default, we adopt Adam \cite{adam} optimizer with a learning rate of 0.0001. 
For training the diffusion network, we employ the following training hyperparameters: step $T$=100, batch size=128 and epochs=400. 
For training $\mathbf{C}_s$, we use batch size=100. 
Note that for the OPA dataset, since it does not provide the semantic segmentation map of each image as the Cityscapes, the $\mathbf{C}_s$ for OPA is directly trained on the composite images instead of the composite structural layouts.
From the comparison results and ablation study, such image-level classifier can also be used to boost the performance for diffusion-based object placement.


\subsection{Evaluation metrics}\label{subsec8}



In our evaluation, we employ multiple metrics to comprehensively assess the quality of the generated composite images in terms of plausibility and diversity.

\textbf{User study on plausibility.} 
To evaluate the plausibility of the composite images, we first conduct a user study in which human participants provide subjective assessments. 
Specifically, the user study was conducted by 20 computer science graduate students. Each questionnaire consisted of 30 image groups, with each group comprising a set of results generated from different methods for placing one foreground object onto a background image. 
Each participant was asked to score each image in every group on a scale of 1-5, based on two criteria: 1) Plausibility of the size and location of the foreground object placed on the background image; 2) Overall structural coherence of the image, independent of factors such as color, shadow and resolution. 

\textbf{Objective metrics on plausibility.}
Additionally, we utilize the accuracy metric to quantitatively evaluate the plausibility of the composite images. 
Following the similar way of defining accuracy in GracoNet \cite{graph}, our accuracy metric is defined as the percentage between the number of positive examples predicted by the plausibility classifier $\mathbf{C}_s$ and the total number of composite images.
Moreover, we employ the FID (Fréchet Inception Distance) metric \cite{fid}, which is also used to measure the realism and plausibility by quantifying the similarity between the distribution of the composite results and that of positive samples in the test set. 

\textbf{Metrics on diversity.}
Furthermore, to assess the diversity and variation in the generated composite images, we adopt the LPIPS (Learned Perceptual Image Patch Similarity) metric \cite{LPIPS}. This perceptual similarity metric captures the dissimilarity between images and serves as an indicator of diversity. Additionally, we measure the ($\Delta{s}, \Delta{h}, \Delta{v}$) values to evaluate the scaling, horizontal, and vertical variations, respectively. Higher values of ($\Delta{s}, \Delta{h}, \Delta{v}$) indicate larger range of object placement variations and higher diversity in the generated composite images. 

\begin{figure*}[t]
\centering
\includegraphics[width=0.98\textwidth]{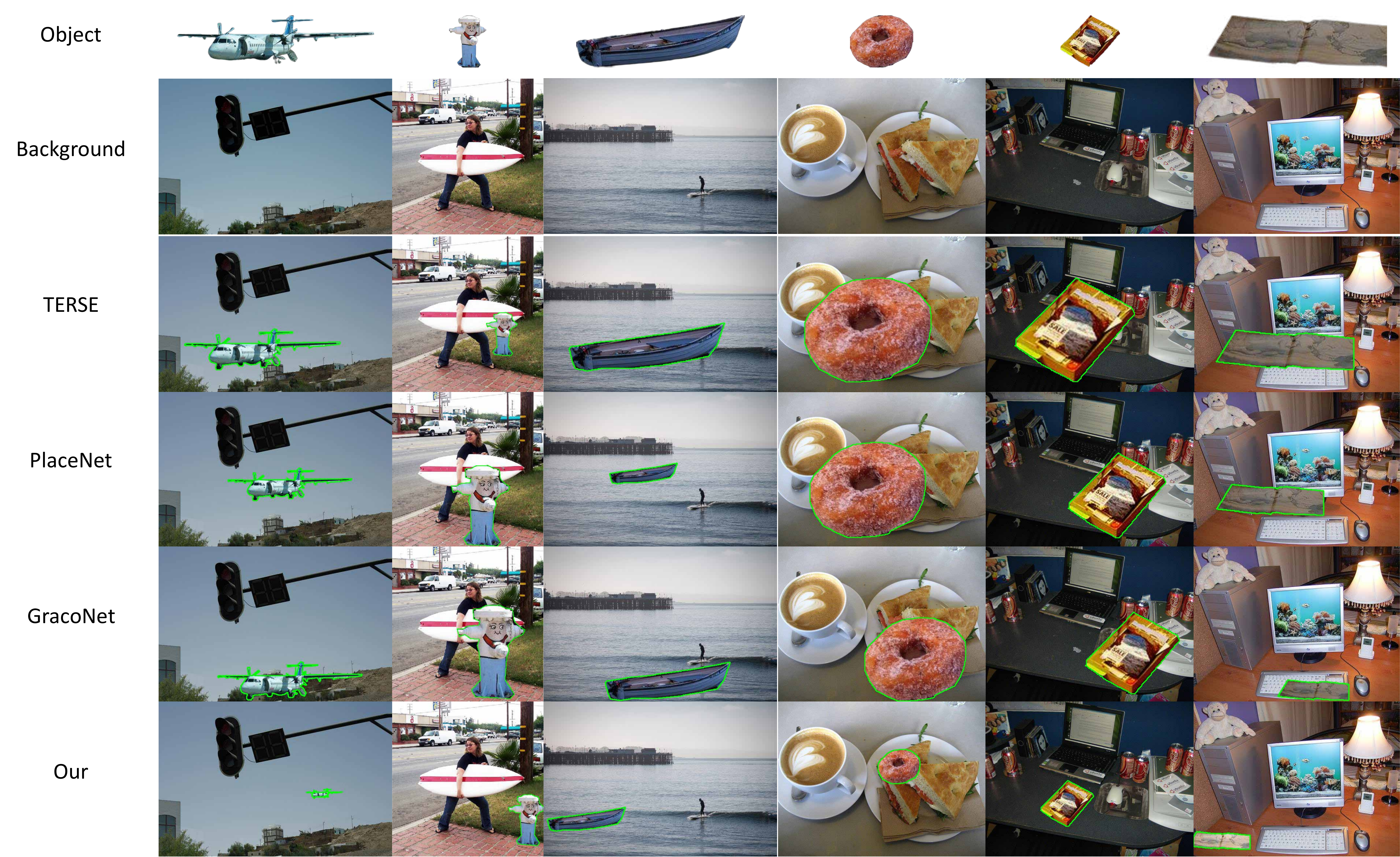} 
\caption{
Qualitative results of single object placement on OPA dataset. Best viewed with zoom-in. 
}
\label{img6}
\end{figure*}

\subsection{Comparisons}\label{subsec9}
For Cityscapes-OP dataset, we compare our DiffPop, against most related methods including TERSE \cite{terse}, PlaceNet \cite{placenet}, GracoNet \cite{graph}, and SAC-GAN \cite{sac}. 
However, for the OPA dataset, the SAC-GAN is not included in the comparison due to the unavailability of semantic segmentation labels. 
For fair comparison, we adopt the experimental setup of GracoNet and incorporate the binary classification loss \cite{graph} into the other methods, enabling their discriminator networks to utilize labeled positive and negative samples. 

\textbf{Qualitative comparisons.}
Fig. \ref{img5} and Fig. \ref{img6} show qualitative comparison results on Cityscapes-OP and OPA datasets. For Cityscapes-OP, we show different representative scene scenarios: empty road, crowded road and pedestrians occupied on the streets etc. 
As shown in Fig. \ref{img5}, our method demonstrates superior performance compared to other methods in these cases.
As our method explicitly consider the structural coherence in the structural plausibility classifier, our results show more plausible structural relations among the inserted objects and other objects in the scene.
Also, comparing the GAN-based methods, our diffusion-based methods can generate object placement with larger diversity, e.g., the vehicle can be placed near the border of the image, instead of mainly appear near the image center. 
Similarly, for OPA dataset, our method outperforms others for different scenes and object categories, by generating object placements with larger variation in object scales and more plausible location that lead to higher scene coherence.

\begin{table}[t]
\large
\centering
\caption{Quantitative object placement results for different methods on Cityscapes-OP dataset. Best results are bolded and second best are underscored.}
\label{tab1}
\resizebox{\columnwidth}{!}{%
\begin{tabular}{c|ccc|cccc}
\multirow{2}{*}{Method} & \multicolumn{3}{c|}{Plausibility}             & \multicolumn{4}{c}{Diversity}                                        \\
                         & User study↑ & Acc.↑          & FID↓           & LPIPS↑         & $\Delta$s↑             & $\Delta$h↑             & $\Delta$v↑             \\ \hline
TERSE                   & ---            & 0.782          & 33.85          &0                &0                 &0                 &0                 \\
PlaceNet                & 2.43            & \underline{0.888}    & 68.50          & \textbf{0.117} & \underline{0.0392}    & \underline{0.0420}    & \textbf{0.0417} \\
GracoNet                 & 2.80            & 0.867          & 34.78          & 0.027          & 0.0199          & 0.0152          & 0.0093          \\
SAC-GAN                 & \underline{2.97}            & 0.822          & \underline{29.98}    & 0.012          & 0.0018          & 0.0180          & 0.0009          \\ \hline
Ours                      & \textbf{3.37}            & \textbf{0.921} & \textbf{18.20} & \underline{0.069}    & \textbf{0.0755} & \textbf{0.2006} & \underline{0.0399}   
\end{tabular}%
}
\end{table}

\begin{table}[t]
\large
\centering
\caption{Quantitative object placement results for different methods on OPA dataset. Best results are bolded and second best are underscored.}
\label{tab2}
\resizebox{\columnwidth}{!}{%
\begin{tabular}{c|ccc|cccc}
\multirow{2}{*}{Method} & \multicolumn{3}{c|}{Plausibility}             & \multicolumn{4}{c}{Diversity}                                        \\
                        & User study↑ & Acc.↑          & FID↓           & LPIPS↑         & $\Delta$s↑             & $\Delta$h↑             & $\Delta$v↑             \\ \hline
TERSE                   & ---            & 0.629          & 46.64          & 0              & 0               & 0               & 0               \\
PlaceNet                & 2.72            & 0.641          & 34.39          & 0.161          & 0.0666          & 0.0756          & \underline{0.0807}          \\
GracoNet               & \underline{2.89}            & \underline{0.798}          & \underline{28.29}          & \underline{0.193}          & \underline{0.1247}          & \underline{0.1302}          & 0.0611          \\ \hline
Ours                     & \textbf{3.59}            & \textbf{0.954} & \textbf{19.76} & \textbf{0.202} & \textbf{0.1536} & \textbf{0.2081} & \textbf{0.1314}
\end{tabular}%
}
\end{table}

\textbf{Quantitative comparisons.}
In Tables \ref{tab1} and \ref{tab2}, we provide quantitative results for object placement and the results demonstrate that our method outperforms the compared methods on both datasets for plausibility measurements. 
For the diversity, since TERSE lacks random noise input, its results exhibit limited diversity on both datasets.
For Cityscapes-OP dataset, since PlaceNet tends to generate larger object scales during for the object placement (see Fig. \ref{img5}), it may result in significant pixel-level changes in the composite images. 
As a result, PlaceNet exhibits a higher LPIPS value, which may not accurately reflect the diversity of the object placement.
However, when considering all the spatial variation metrics ($\Delta{s}, \Delta{h}, \Delta{v}$) that reflect the diversity of generated object scales and locations, our method outperforms PlaceNet. 
Therefore, developing a more comprehensive evaluation metric of diversity, e.g., taking into account both LPIPS and ($\Delta{s}, \Delta{h}, \Delta{v}$) metrics, will be worthy to investigate in the future.

\begin{table}[t]
\centering
\caption{Quantitative results of data augmentation with different methods on Cityscapes dataset. IoU ($\uparrow$) for evaluating image segmentation performance for different categories are compared. Note the data augmentation is only conducted for the Car category.}
\label{tab3}
\resizebox{\columnwidth}{!}{%
\begin{tabular}{c|cccc|cccc}
\multirow{2}{*}{Data} & \multicolumn{4}{c|}{PSPNet}                                       & \multicolumn{4}{c}{DeepLabv3}                                     \\
                      & Car         & Truck       & Bus         & mIoU           & Car         & Truck       & Bus         & mIoU           \\ \hline
Original              & 93.12          & 61.66          & 77.07          & 70.73          & 93.99          & 73.22          & 78.28          & 73.50          \\
+TERSE                & 93.28          & 57.32          & 72.38          & 69.77          & 94.30          & \textbf{76.41} & 82.31          & 73.95          \\
+PlaceNet            & 93.15          & 61.61          & 69.06          & 68.14          & 93.72          & 65.91          & 73.81          & 71.82          \\
+GracoNet            & 93.38          & 63.98          & 73.19          & 69.50          & 93.84          & 69.84          & 77.16          & 72.87          \\
+SAC-GAN              & 93.17          & 56.73          & 73.08          & 70.29          & 94.25          & 74.33          & 81.29          & 73.84          \\ \hline
+Ours                   & \textbf{93.42} & \textbf{64.56} & \textbf{77.97} & \textbf{71.11} & \textbf{94.34} & 75.28          & \textbf{82.70} & \textbf{74.75}
\end{tabular}%
}
\end{table}

\subsection{Applications}\label{subsec10}
\textbf{Data augmentation.}
We also investigate the potential of our approach for data augmentation, specifically for semantic segmentation tasks. 
First, we randomly select 100 car objects from the total object library of the Cityscapes-OP dataset. 
Then, for each image in the training set of the Cityscapes dataset, we randomly select a car object as the foreground and place it onto the background.
For each compared object placement method, we generate 2,975 composite images for data augmentation. Subsequently, we train two semantic segmentation models PSPNet \cite{pspnet} and DeepLabv3 \cite{deeplabv3}, on the augmented data generated by different methods. 
To evaluate the performance of each method in terms of data augmentation, we employ the Intersection-over-Union (IoU) metrics to quantize the segmentation performance on individual object classes and compare the results of models trained with augmented images.
The results in Table \ref{tab3} demonstrate that our method outperforms baselines in most cases. It is important to note that in this experiment, the data augmentation is only performed for the Car class, and the expected outcome is to increase the IoU for Car, without compromising the IoU for other categories.
Meanwhile, it is interesting to see that the performance Truck and Bus can also be consistently increased, which may be due to the less confusion among the three categories.
Such results can verify the plausibility and diversity of our generated composite images to a certain degree.

\begin{figure*}[htbp]
\centering

\includegraphics[width=\textwidth]{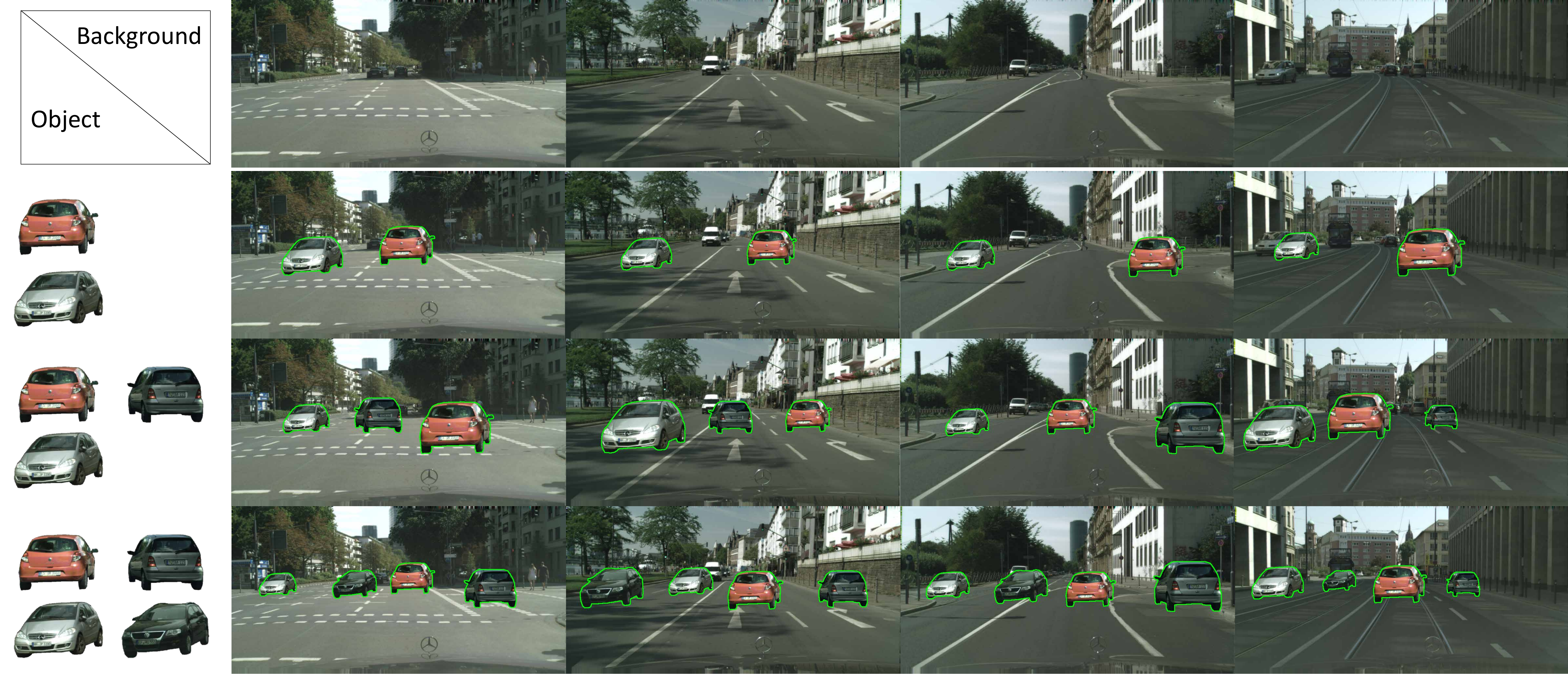} 

\caption{
Qualitative results of multi-object placement on Cityscapes-OP. Best viewed with zoom-in. 
}
\label{img7}
\end{figure*}

\textbf{Multi-object placement.}
Our DiffPop framework can also be extended for multi-object placement.
Similar to the structural plausibility classifier, a \textit{relational} plausibility classifier $\mathbf{C}_r$ can be trained to guide the diffusion sampling so that the generated results contain plausible spatial relationships between multiple objects.
The relational plausibility classifier $\mathbf{C}_r$ is trained to discriminate whether the relation between two independently generated object placements are plausible or not.
More details about how to train $\mathbf{C}_r$ and how to use it together with the structural plausibility classifier $\mathbf{C}_s$ in the guided diffusion process are provided in the supplementary material.
To evaluate the performance of our method for multi-object placement, we conduct preliminary qualitative experiments on Cityscapes-OP, as shown in Figure \ref{img7}. Specifically, we selected 4 objects from the object library and 4 backgrounds from the background library. Then, we separately place 2, 3, and 4 objects on each of the four backgrounds, respectively. 
The results demonstrate that our method can be effectively extended for multi-object placement.
More exploration and evaluation of our diffusion-based framework for multi-object placement will also be a promising future direction.

\begin{table}[]
\centering
\caption{Quantitative results of ablation study on $\mathbf{C}_s$ and guidance scale $\lambda$ on Cityscapes-OP dataset.}
\label{tab4}
\resizebox{0.9\columnwidth}{!}{%
\begin{tabular}{c|cc|cccc}
\multirow{2}{*}{$\lambda$} & \multicolumn{2}{c|}{Plausibility} & \multicolumn{4}{c}{Diversity}     \\
                               & Acc.↑           & FID↓            & LPIPS↑ & $\Delta$s↑    & $\Delta$h↑    & $\Delta$v↑    \\ \hline
0                              & 0.558           & 20.42           & 0.059  & 0.0757 & 0.2119 & 0.0425 \\
0.001                          & 0.756           & 19.23           & 0.062  & 0.0752 & 0.2085 & 0.0404 \\
0.002                          & 0.816           & 18.91           & 0.063  & 0.0756 & 0.2049 & 0.0394 \\
0.005                          & 0.907           & 18.48           & 0.065  & 0.0739 & 0.2021 & 0.0380 \\ \hline
0.01                           & 0.921           & 18.20           & 0.069  & 0.0755 & 0.2006 & 0.0399 \\ \hline
0.02                           & 0.912           & 18.66           & 0.075  & 0.0794 & 0.2046 & 0.0473 \\
0.05                           & 0.809           & 20.53           & 0.087  & 0.1017 & 0.2050 & 0.0801 \\
0.1                            & 0.736           & 22.54           & 0.101  & 0.1223 & 0.2100 & 0.1154 \\
0.2                            & 0.759           & 24.36           & 0.111  & 0.1297 & 0.1982 & 0.1268 \\
0.5                            & 0.852           & 25.64           & 0.115  & 0.1239 & 0.1876 & 0.1008
\end{tabular}%
}
\end{table}

\begin{table}[]
\centering
\caption{Quantitative results of ablation study on $\mathbf{C}_s$ and guidance scale $\lambda$ on OPA dataset.}
\label{tab5}
\resizebox{0.9\columnwidth}{!}{%
\begin{tabular}{c|cc|cccc}
\multirow{2}{*}{$\lambda$} & \multicolumn{2}{c|}{Plausibility} & \multicolumn{4}{c}{Diversity}     \\
                               & Acc.↑           & FID↓            & LPIPS↑ & $\Delta$s↑    & $\Delta$h↑    & $\Delta$v↑    \\ \hline
0                              & 0.531           & 22.04           & 0.175  & 0.1829 & 0.1964 & 0.1809 \\
0.01                           & 0.823           & 19.82           & 0.177  & 0.1783 & 0.2292 & 0.1876 \\
0.02                           & 0.886           & 19.46           & 0.183  & 0.1709 & 0.2269 & 0.1714 \\
0.05                           & 0.933           & 19.71           & 0.196  & 0.1609 & 0.2159 & 0.1466 \\
0.1                            & 0.947           & 20.11           & 0.202  & 0.1565 & 0.2082 & 0.1362 \\ \hline
0.2                            & 0.954           & 19.76           & 0.202  & 0.1536 & 0.2081 & 0.1314 \\ \hline
0.5                            & 0.947           & 19.19           & 0.199  & 0.1545 & 0.2146 & 0.1350 \\
1                              & 0.925           & 18.71           & 0.197  & 0.1660 & 0.2281 & 0.1522 \\
2                              & 0.841           & 18.39           & 0.185  & 0.2076 & 0.2645 & 0.2171 \\
5                              & 0.610           & 25.04           & 0.130   & 0.2540 & 0.2881 & 0.3012
\end{tabular}%
}
\end{table}

\subsection{Ablation study}
\label{sec:ablation}
\textbf{Guidance scale $\lambda$.}
We conducted experiments to evaluate the effectiveness of $\mathbf{C}_s$ and the impact of different guidance scale factors $\lambda$ on the performance of our method. The results are presented in Table~\ref{tab4} and Table \ref{tab5}, where a guidance scale of $0$ indicates that $\mathbf{C}_s$ was not used for guidance. 
It can be seen that the guidance provided by $\mathbf{C}_s$ has a significant impact on all the evaluated metrics. When an appropriate guidance scale factor is used, notable improvements in accuracy, FID, and LPIPS can be obtained. However, if the guidance scale $\lambda$ is too large, the accuracy and FID metrics tend to decrease, while the ($\Delta{s}, \Delta{h}, \Delta{v}$) metrics increase. 
This is due to excessive guidance applied to the sampling of the diffusion model, which may lead to the generation of scales and locations that deviate from the real distribution. 
Notably, the selection of appropriate $\lambda$ differs for different datasets. 
The OPA dataset contains multiple foreground and background classes, resulting in a wider distribution of real scales and locations. 
As a result, OPA exhibits a stronger tolerance to larger guidance scale factors.
In contrast, the real object scales and locations in Cityscapes-OP dataset are relatively limited comparing to OPA.
Hence, the $\lambda$ values are generally smaller than those of OPA's, which means weaker guidance is tolerated for Cityscapes-OP.
Based on this experiment, we use $\lambda=0.01$ for Cityscapes-OP and $\lambda=0.2$ for all other experiments in this paper.


\begin{table}[]
\centering
\caption{Quantitative results of ablation study on influence of different input type of $\mathbf{C}_s$ on Cityscapes-OP dataset.}
\label{tab6}
\resizebox{\columnwidth}{!}{%
\begin{tabular}{c|c|ccc}
Input Type                    & F1↑ & TPR↑   & TNR↑   & Balanced   Acc↑ \\ \hline
Mask \hangz{+}   composite \hangz{i}mage     & 0.709    & 0.734 & 0.780 & 0.757               \\
Mask \hangz{+}    \hangz{semantic label} & 0.797    & 0.838 & 0.826 & 0.832               \\ \hline
Mask \hangz{+}    \hangz{binarized layout}       & 0.864    & 0.906 & 0.874 & 0.890              
\end{tabular}%
}
\end{table}

\textbf{Input of the structural plausibility classifier $\mathbf{C}_s$.}
We conducted experiments on Cityscapes-OP to assess the effect of training $\mathbf{C}_s$ on different types of inputs:
RGB image with object mask, semantic label with object mask, and binarized semantic layout with object mask.
Here, the binarized semantic layout is created by converting each semantic label to a separate binary mask and then concatenating them into a binarized semantic layout with 19 channels.
From the results in Table \ref{tab6}, it can be observed the binarized semantic layout with object mask as the input for $\mathbf{C}_s$ yields the best performance in terms of F1-score, true positive rate (TPR), true negative rate (TNR) and balanced accuracy for evaluating the plausibility (defined in a similar way to the accuracy in Sec. \ref{subsec8}).
The reason may be the binarized semantic layout can capture more disentangled scene structure and richer spatial features compared to the original images and  semantic labels.


\begin{figure}[htbp]
\centering

\includegraphics[width=\columnwidth]{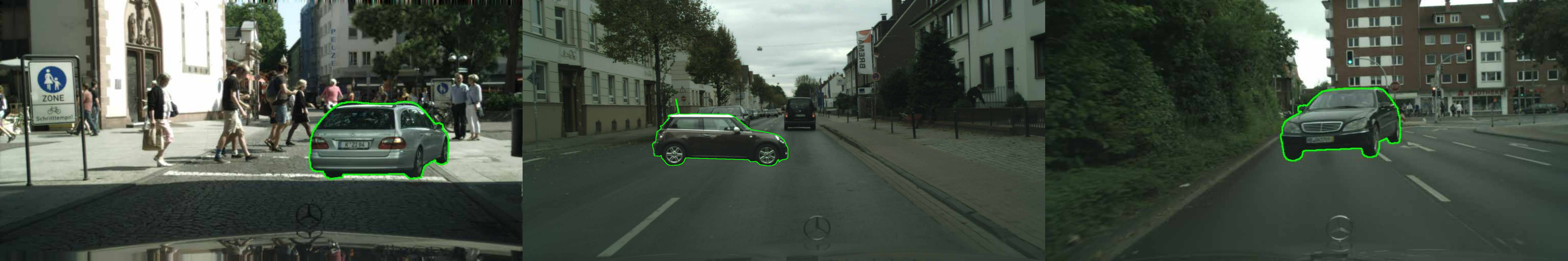} 

\caption{Failure cases on specific scenarios. Best viewed with zoom-in.}

\label{img8}
\end{figure}

\subsection{Limitations}
Though our method achieves promising results for plausible object placement, it still has certain limitations when dealing with specific scenarios, including cases with complex backgrounds, mismatched foreground and background, and adherence to traffic rules. Firstly, the complexity of backgrounds can pose challenges for generating realistic composite images, as observed in the first sample of Fig. \ref{img8}. 
Secondly, our method struggles to handle situations where the foreground and background do not match, as illustrated in the second example. For example, if the foreground object depicts a car facing left and right while the background represents a road facing up and down, our method encounters difficulties in determining the compatibility between them, resulting in the generation of unrealistic composite images. 
Lastly, our approach currently lacks the ability to enforce adherence to traffic rules, as shown in the third example.
Without the knowledge of different types of roads and the more detailed semantic information, it is hard to ensure proper placement of vehicles in accordance with specific traffic regulations.

%% file: 05-conclusion.tex
\section{Conclusion}

We present DiffPop, a novel framework to learn object placement via plausibility-guided diffusion model. 
In contrast to previous works, our approach achieves superior performance in generating more plausible and diverse object placement, as well as more robust training comparing to the GAN-based methods.
Specifically, our approach leverages a structural plausibility classifier to guide the diffusion model in sampling more reasonable scales and locations for  the given object.
To train the plausibility classifier, we introduce the Cityscapes-OP, a new dataset annotated with positive and negative plausibility labels, which can facilitate future research endeavors on plausible object placement learning.
Besides, our DiffPop framework is also extensible to multi-object placement, showcasing its potential and efficacy in more complex image composition tasks. 
In the future, one interesting direction is to enhancing the object placement learning so that it can generalize to unseen categories, without intensive training.
In addition, it is also worthy to explore how to integrate the object placement with the image harmonization into the same diffusion-based framework to create realistic image composition.

%% file: 06-Supplementary-material.tex
\section{Supplementary Material}
In this supplementary material, we provide additional details on the creation of Cityscapes-OP dataset and more technical details on how to extend our framework for multi-object placement.

\subsection{More details on Cityscapes-OP dataset}
The Cityscapes \cite{cityscapes} dataset, initially designed for 2D semantic segmentation of street scenes, consists of 2,975 training and 500 validation images. Due to its complex background scenes, this dataset has been widely used in previous object placement methods. 
To enable training of the plausibility classifier which needs both positive and negative samples
we process the Cityscapes dataset to create a dataset specifically for object placement, referred to as Cityscapes-OP. 
When constructing the Cityscapes-OP dataset, we first extract 1,300 intact cars from the Cityscapes dataset to form the total object library. From this library, we randomly select 50 cars for the training object library and 25 cars for the test object library. Additionally, we randomly choose 500 images from the Cityscapes training set and 100 images from the validation set to create the training background library and test background library, respectively. Given the complexity of the background scenes, our focus in building the dataset lies primarily on the background images rather than the foreground objects.

To build the training set, we utilize our pre-trained unguided diffusion model as the generator to produce 10,000 sizes and locations for object placement. By randomly composing the one of 50 cars from the training object library with the one of the 500 background images from the training background library, we obtain 10,000 composite images. These images are then manually labeled as plausible or implausible, resulting in 3,869 positive and 6,131 negative samples. To further enrich the dataset, we include 2,000 additional simple negative samples, such as composite images with objects that are over large or over small, or placed in unreasonable locations. These simple negative samples are created without overlapping with the previous 10,000 samples in terms of objects and backgrounds. Finally, the labeled 12,000 composite images constitute the training set of the Cityscapes-OP dataset. The test set is constructed in a similar way to the training set, where we generate 1,000 sizes and locations to create composite images by randomly combining one of the 25 objects from the test object library with one of the 100 backgrounds from the test background library. The resulting 1,000 composite images are manually labeled, yielding 395 positive samples and 605 negative samples.

It is worthy noting that the Cityscapes-OP dataset primarily focuses on single foreground category, specifically for cars. In contrast, the OPA dataset encompasses multiple foreground classes. On the other hand, the complex background scenes of Cityscapes-OP dataset allow us to effectively evaluate object placement methods in challenging background scenarios. During the dataset construction, we leverage the annotated bounding boxes and semantic/instance segmentation in the Cityscapes to create the Cityscapes-OP. These annotations provide valuable information for the foreground object generation and enhance the quality and realism of the composite images. Taking these factors into consideration, we believe the Cityscapes-OP dataset will be valuable for advancing research in the field of object placement.

\begin{figure}[t]
\centering

\includegraphics[width=0.49\textwidth]{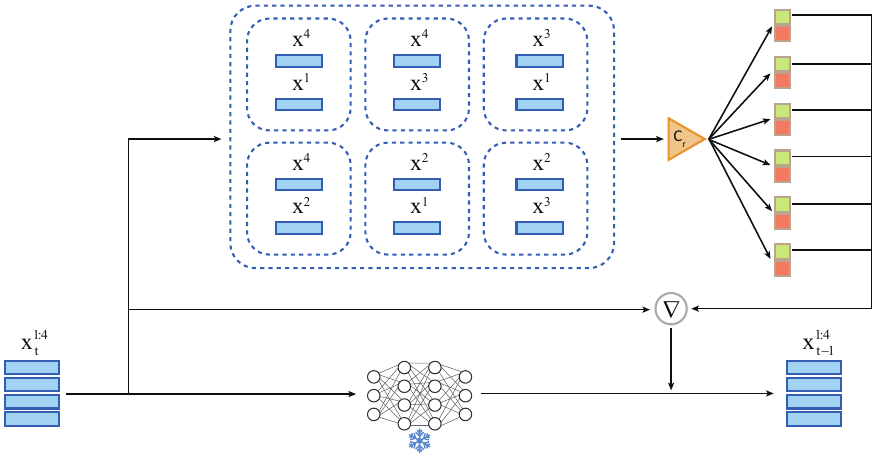} 
\caption{
The details of $\mathbf{C}_r$ during inference. Taking the example of placing 4 objects, we first pair up the locations of the 4 objects obtained from the diffusion model sampling at time step t, resulting in 6 location pairs. Then, the $\mathbf{C}_r$ guides the sampling results of the diffusion model at time step $t-1$ based on the relational plausibility score of these 6 location pairs.
}
\label{multi-detail}
\end{figure}

\begin{figure*}[ht]
\centering
\includegraphics[width=\textwidth]{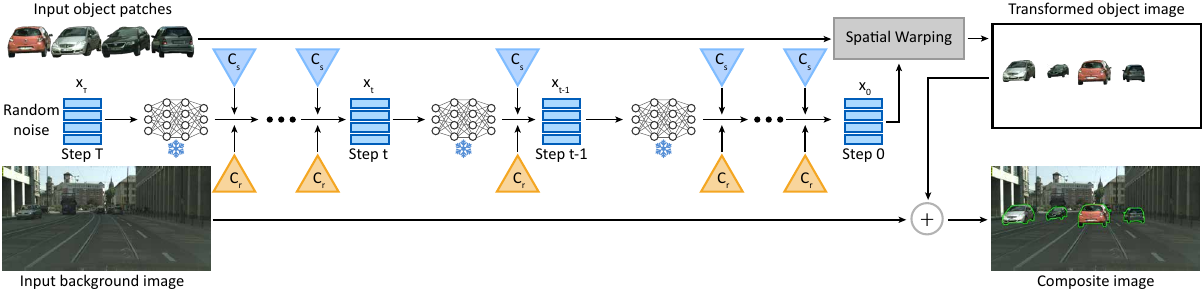} 
\caption{ Illustration of how to use
$\mathbf{C}_r$ and $\mathbf{C}_s$ together to guide the diffusion model sampling. 
}
\label{multi_total}
\end{figure*}

\subsection{More details on relational plausibility classifier}

We propose a relational plausibility classifier $\mathbf{C}_{{r}}$ to guide the diffusion model sampling so that the sampling direction is biased towards the direction of plausible spatial relationship between multiple objects, enabling the diffusion model to generate appropriate sizes and locations for multiple foreground objects.

The network structure of $\mathbf{C}_{{r}}$ consists of four fully connected layers. Its goal is to determine whether the spatial relationship between two objects is plausible or not. $\mathbf{C}_{{r}}$ takes the placements of two objects as input and outputs a two-dimensional vector representing the plausibility of their spatial relationship.

$\mathbf{C}_{{r}}$ is trained in a supervised manner with cross-entropy loss function. 
The training data are also manually labeled based on the Cityscapes-OP dataset. 
Firstly, 5 objects are randomly selected from the training object library of the Cityscapes-OP dataset, and 80 background images are randomly selected from the training background library. 
Then, each 2 of these 5 objects is paired together, resulting in 10 pairs of objects. For each pair of objects, image composition is performed under the same background using the $\mathbf{C}_{{s}}$ guided diffusion sampling, yielding composite images with two objects placed within the scene. 
Thereby, 800 composite images are obtained. These images are then manually labeled, following the same annotation process as the Cityscapes-OP dataset, and result in 407 positive samples and 393 negative samples. The 800 labeled composite images are then used to train $\mathbf{C}_{{r}}$. This training is conducted on an RTX 3060 GPU using the Adam optimizer, with a batch size of 100 and 400 training epochs at a learning rate of $10^{-4}$.

In the case of one background and two foreground objects, the noisy placements $ \mathbf{x}_{{t}}^{1} $, $ \mathbf{x}_{{t}}^{2} $ for the two objects sampled by the diffusion model at step $ t $ are fed into the relational plausibility classifier $\mathbf{C}_{{r}}$. The gradient of the classifier's output with respect to its input is then computed. This gradient is used to correct the mean of the distribution predicted by the diffusion model, thereby steering the sampling direction of the diffusion model towards a more plausible spatial relationship between the two objects. The sampling equation is as follows:

\begin{equation}
\label{eq:1}
\begin{aligned}
    (\mathbf{x}_{{t}-1}^{1}, \mathbf{x}_{{t}-1}^{2}) \leftarrow {N}\left(\mu+\lambda_{{r}} \Sigma \nabla_{\left(\mathbf{x}_{{t}}^{1}, \mathbf{x}_{{t}}^{2}\right)} \log {p}_{\tau}(\mathbf{y} \mid \mathbf{x}_{{t}}^{1}, \mathbf{x}_{{t}}^{2}), \Sigma\right),
\end{aligned}
\end{equation}
where $\lambda_{{r}}$ is the guidance scale factor, used to control the degree of distribution correction. $\Sigma$ is a fixed constant obtained from the diffusion process calculation.

Fig. \ref{multi-detail} shows the case of taking 4 objects for multi-object placement.
Specifically, the diffusion model samples four noisy placements at step $ t $: $ \mathbf{x}_{{t}}^{1} $, $ \mathbf{x}_{{t}}^{2} $, $ \mathbf{x}_{{t}}^{3} $, $ \mathbf{x}_{{t}}^{4} $. 
These placements are paired to obtain six placement combinations. Then, for each combination, the gradient is computed by Eq.~(\ref{eq:1}). 
The gradients of all combinations are summed and used to correct the mean $ \mu $ of the distribution predicted by the diffusion model.

Fig. \ref{multi_total} shows how to use the $\mathbf{C}_{{r}}$ together with the $\mathbf{C}_{{s}}$ to guide the sampling of the diffusion model, thereby accomplishing the task of placing multiple foreground objects on a single background image. The $\mathbf{C}_{{s}}$ ensures that the size and location of each object are plausible, while the $\mathbf{C}_{{r}}$ ensures that the spatial relationships between the objects are plausible.

%% file: main.bbl
\newcommand{\etalchar}[1]{$^{#1}$}
\begin{thebibliography}{\uppercase{GPAM{\etalchar{*}}20}}

\bibitem[COR{\etalchar{*}}16]{cityscapes}
\textsc{Cordts M., Omran M., Ramos S., Rehfeld T., Enzweiler M., Benenson R., Franke U., Roth S., Schiele B.}:
\newblock The {C}ityscapes dataset for semantic urban scene understanding.
\newblock In \emph{Proceedings of the IEEE Conference on Computer Vision and Pattern Recognition} (2016), pp.~3213--3223.

\bibitem[CPSA17]{deeplabv3}
\textsc{Chen L.-C., Papandreou G., Schroff F., Adam H.}:
\newblock Rethinking atrous convolution for semantic image segmentation.
\newblock \emph{arXiv preprint arXiv:1706.05587} (2017).

\bibitem[DN21]{beat-gan}
\textsc{Dhariwal P., Nichol A.}:
\newblock Diffusion models beat {GANs} on image synthesis.
\newblock \emph{Advances in Neural Information Processing Systems 34} (2021), 8780--8794.

\bibitem[FSW{\etalchar{*}}19]{o3}
\textsc{Fang H.-S., Sun J., Wang R., Gou M., Li Y.-L., Lu C.}:
\newblock Instaboost: Boosting instance segmentation via probability map guided copy-pasting.
\newblock In \emph{Proceedings of the IEEE/CVF International Conference on Computer Vision} (2019), pp.~682--691.

\bibitem[GMBK17]{o4}
\textsc{Georgakis G., Mousavian A., Berg A.~C., Kosecka J.}:
\newblock Synthesizing training data for object detection in indoor scenes.
\newblock \emph{arXiv preprint arXiv:1702.07836} (2017).

\bibitem[GPAM{\etalchar{*}}20]{GAN}
\textsc{Goodfellow I., Pouget-Abadie J., Mirza M., Xu B., Warde-Farley D., Ozair S., Courville A., Bengio Y.}:
\newblock Generative adversarial networks.
\newblock \emph{Communications of the ACM 63}, 11 (2020), 139--144.

\bibitem[HJA20]{ddpm}
\textsc{Ho J., Jain A., Abbeel P.}:
\newblock Denoising diffusion probabilistic models.
\newblock \emph{Advances in Neural Information Processing Systems 33} (2020), 6840--6851.

\bibitem[HRU{\etalchar{*}}17]{fid}
\textsc{Heusel M., Ramsauer H., Unterthiner T., Nessler B., Hochreiter S.}:
\newblock {GANs} trained by a two time-scale update rule converge to a local nash equilibrium.
\newblock \emph{Advances in Neural Information Processing Systems 30} (2017).

\bibitem[HS22]{class-free}
\textsc{Ho J., Salimans T.}:
\newblock Classifier-free diffusion guidance.
\newblock \emph{arXiv preprint arXiv:2207.12598} (2022).

\bibitem[HZO{\etalchar{*}}23]{image-compos-diffuion}
\textsc{Hachnochi R., Zhao M., Orzech N., Gal R., Mahdavi-Amiri A., Cohen-Or D., Bermano A.~H.}:
\newblock Cross-domain compositing with pretrained diffusion models.
\newblock \emph{arXiv preprint arXiv:2302.10167} (2023).

\bibitem[HZRS16]{resnet}
\textsc{He K., Zhang X., Ren S., Sun J.}:
\newblock Deep residual learning for image recognition.
\newblock In \emph{Proceedings of the IEEE Conference on Computer Vision and Pattern Recognition} (2016), pp.~770--778.

\bibitem[JSZ{\etalchar{*}}15]{stn}
\textsc{Jaderberg M., Simonyan K., Zisserman A., et~al.}:
\newblock Spatial transformer networks.
\newblock \emph{Advances in Neural Information Processing Systems 28} (2015).

\bibitem[KB14]{adam}
\textsc{Kingma D.~P., Ba J.}:
\newblock Adam: A method for stochastic optimization.
\newblock \emph{arXiv preprint arXiv:1412.6980} (2014).

\bibitem[KMR{\etalchar{*}}23]{sam}
\textsc{Kirillov A., Mintun E., Ravi N., Mao H., Rolland C., Gustafson L., Xiao T., Whitehead S., Berg A.~C., Lo W.-Y., et~al.}:
\newblock Segment anything.
\newblock \emph{arXiv preprint arXiv:2304.02643} (2023).

\bibitem[LLG{\etalchar{*}}18]{context}
\textsc{Lee D., Liu S., Gu J., Liu M.-Y., Yang M.-H., Kautz J.}:
\newblock Context-aware synthesis and placement of object instances.
\newblock \emph{Advances in Neural Information Processing Systems 31} (2018).

\bibitem[LLZ{\etalchar{*}}21]{opa}
\textsc{Liu L., Liu Z., Zhang B., Li J., Niu L., Liu Q., Zhang L.}:
\newblock {OPA}: Object placement assessment dataset.
\newblock \emph{arXiv preprint arXiv:2107.01889} (2021).

\bibitem[LMB{\etalchar{*}}14]{coco}
\textsc{Lin T.-Y., Maire M., Belongie S., Hays J., Perona P., Ramanan D., Doll{\'a}r P., Zitnick C.~L.}:
\newblock Microsoft coco: Common objects in context.
\newblock In \emph{Computer Vision--ECCV 2014: 13th European Conference, Zurich, Switzerland, September 6-12, 2014, Proceedings, Part V 13} (2014), Springer, pp.~740--755.

\bibitem[LPA{\etalchar{*}}23]{more-control}
\textsc{Liu X., Park D.~H., Azadi S., Zhang G., Chopikyan A., Hu Y., Shi H., Rohrbach A., Darrell T.}:
\newblock More control for free! {Image} synthesis with semantic diffusion guidance.
\newblock In \emph{Proceedings of the IEEE/CVF Winter Conference on Applications of Computer Vision} (2023), pp.~289--299.

\bibitem[LSLW16]{vae-gan}
\textsc{Larsen A. B.~L., S{\o}nderby S.~K., Larochelle H., Winther O.}:
\newblock Autoencoding beyond pixels using a learned similarity metric.
\newblock In \emph{International Conference on Machine Learning} (2016), PMLR, pp.~1558--1566.

\bibitem[LYW{\etalchar{*}}18]{st-gan}
\textsc{Lin C.-H., Yumer E., Wang O., Shechtman E., Lucey S.}:
\newblock {ST-GAN}: Spatial transformer generative adversarial networks for image compositing.
\newblock In \emph{Proceedings of the IEEE Conference on Computer Vision and Pattern Recognition} (2018), pp.~9455--9464.

\bibitem[NCL{\etalchar{*}}21]{survey}
\textsc{Niu L., Cong W., Liu L., Hong Y., Zhang B., Liang J., Zhang L.}:
\newblock Making images real again: A comprehensive survey on deep image composition.
\newblock \emph{arXiv preprint arXiv:2106.14490} (2021).

\bibitem[ND21]{iddpm}
\textsc{Nichol A.~Q., Dhariwal P.}:
\newblock Improved denoising diffusion probabilistic models.
\newblock In \emph{International Conference on Machine Learning} (2021), PMLR, pp.~8162--8171.

\bibitem[NDR{\etalchar{*}}21]{glide}
\textsc{Nichol A., Dhariwal P., Ramesh A., Shyam P., Mishkin P., McGrew B., Sutskever I., Chen M.}:
\newblock {GLIDE}: Towards photorealistic image generation and editing with text-guided diffusion models.
\newblock \emph{arXiv preprint arXiv:2112.10741} (2021).

\bibitem[PGM{\etalchar{*}}19]{pytorch}
\textsc{Paszke A., Gross S., Massa F., Lerer A., Bradbury J., Chanan G., Killeen T., Lin Z., Gimelshein N., Antiga L., Desmaison A., Kopf A., Yang E., DeVito Z., Raison M., Tejani A., Chilamkurthy S., Steiner B., Fang L., Bai J., Chintala S.}:
\newblock Pytorch: An imperative style, high-performance deep learning library.
\newblock In \emph{Advances in Neural Information Processing Systems 32}. 2019, pp.~8024--8035.

\bibitem[RBL{\etalchar{*}}22]{stabel-diffusion}
\textsc{Rombach R., Blattmann A., Lorenz D., Esser P., Ommer B.}:
\newblock High-resolution image synthesis with latent diffusion models.
\newblock In \emph{Proceedings of the IEEE/CVF Conference on Computer Vision and Pattern Recognition} (2022), pp.~10684--10695.

\bibitem[RDN{\etalchar{*}}22]{dalle2}
\textsc{Ramesh A., Dhariwal P., Nichol A., Chu C., Chen M.}:
\newblock Hierarchical text-conditional image generation with {CLIP} latents.
\newblock \emph{arXiv preprint arXiv:2204.06125} (2022).

\bibitem[RHB18]{o1}
\textsc{Remez T., Huang J., Brown M.}:
\newblock Learning to segment via cut-and-paste.
\newblock In \emph{Proceedings of the European Conference on Computer Vision} (2018), pp.~37--52.

\bibitem[SCS{\etalchar{*}}22]{imagen}
\textsc{Saharia C., Chan W., Saxena S., Li L., Whang J., Denton E.~L., Ghasemipour K., Gontijo~Lopes R., Karagol~Ayan B., Salimans T., et~al.}:
\newblock Photorealistic text-to-image diffusion models with deep language understanding.
\newblock \emph{Advances in Neural Information Processing Systems 35} (2022), 36479--36494.

\bibitem[SME20]{ddim}
\textsc{Song J., Meng C., Ermon S.}:
\newblock Denoising diffusion implicit models.
\newblock \emph{arXiv preprint arXiv:2010.02502} (2020).

\bibitem[TCA{\etalchar{*}}19]{terse}
\textsc{Tripathi S., Chandra S., Agrawal A., Tyagi A., Rehg J.~M., Chari V.}:
\newblock Learning to generate synthetic data via compositing.
\newblock In \emph{Proceedings of the IEEE/CVF Conference on Computer Vision and Pattern Recognition} (2019), pp.~461--470.

\bibitem[WWY{\etalchar{*}}19]{o2}
\textsc{Wang H., Wang Q., Yang F., Zhang W., Zuo W.}:
\newblock Data augmentation for object detection via progressive and selective instance-switching.
\newblock \emph{arXiv preprint arXiv:1906.00358} (2019).

\bibitem[ZIE{\etalchar{*}}18]{LPIPS}
\textsc{Zhang R., Isola P., Efros A.~A., Shechtman E., Wang O.}:
\newblock The unreasonable effectiveness of deep features as a perceptual metric.
\newblock In \emph{Proceedings of the IEEE Conference on Computer Vision and Pattern Recognition} (2018), pp.~586--595.

\bibitem[ZLC{\etalchar{*}}23]{topnet}
\textsc{Zhu S., Lin Z., Cohen S., Kuen J., Zhang Z., Chen C.}:
\newblock {TopNet}: Transformer-based object placement network for image compositing.
\newblock \emph{arXiv preprint arXiv:2304.03372} (2023).

\bibitem[ZLNZ22]{graph}
\textsc{Zhou S., Liu L., Niu L., Zhang L.}:
\newblock Learning object placement via dual-path graph completion.
\newblock In \emph{Proceedings of the European Conference on Computer Vision} (2022), Springer, pp.~373--389.

\bibitem[ZMZ{\etalchar{*}}22]{sac}
\textsc{Zhou H., Ma R., Zhang L.-X., Gao L., Mahdavi-Amiri A., Zhang H.}:
\newblock {SAC-GAN}: Structure-aware image composition.
\newblock \emph{IEEE Transactions on Visualization and Computer Graphics} (2022).

\bibitem[ZSQ{\etalchar{*}}17]{pspnet}
\textsc{Zhao H., Shi J., Qi X., Wang X., Jia J.}:
\newblock Pyramid scene parsing network.
\newblock In \emph{Proceedings of the IEEE Conference on Computer Vision and Pattern Recognition} (2017), pp.~2881--2890.

\bibitem[ZWM{\etalchar{*}}20]{placenet}
\textsc{Zhang L., Wen T., Min J., Wang J., Han D., Shi J.}:
\newblock Learning object placement by inpainting for compositional data augmentation.
\newblock In \emph{Proceedings of the European Conference on Computer Vision} (2020), Springer, pp.~566--581.

\bibitem[ZZL{\etalchar{*}}20]{o5}
\textsc{Zhang S.-H., Zhou Z.-P., Liu B., Dong X., Hall P.}:
\newblock What and where: A context-based recommendation system for object insertion.
\newblock \emph{Computational Visual Media 6} (2020), 79--93.

\end{thebibliography}
